# Features Fusion for Classification of Logos

N. Vinay Kumar[a]*, Pratheek[a], V. Vijaya Kantha[a], K. N. Govindaraju[a], and D. S. Guru[a]

[a]*Department. of Studies in Computer Science, University of Mysore, Mysore-570006, INDIA*

**Abstract**

In this paper, a logo classification system based on the appearance of logo images is proposed. The proposed classification system makes use of global characteristics of logo images for classification. Color, texture, and shape of a logo wholly describe the global characteristics of logo images. The various combinations of these characteristics are used for classification. The combination contains only with single feature or with fusion of two features or fusion of all three features considered at a time respectively. Further, the system categorizes the logo image into: a logo image with fully text or with fully symbols or containing both symbols and texts.. The K-Nearest Neighbour (K-NN) classifier is used for classification. Due to the lack of color logo image dataset in the literature, the same is created consisting 5044 color logo images. Finally, the performance of the classification system is evaluated through accuracy, precision, recall and F-measure computed from the confusion matrix. The experimental results show that the most promising results are obtained for fusion of features.

*Keywords:* Appearance based features, Feature level fusion, Logo image classification,

## 1. Introduction

With the rapid development of multimedia information technology, the amount of image data available on the internet is very huge and it is increasing exponentially. Handling of such a huge quantity of image data has become a more challenging and at the same time it is an interesting research problem. Nowadays, to handle such image data, there are lot many tools available on the internet such as ArcGIS, Google, Yahoo, Bing etc. Currently, those tools perform classification, detection, and retrieval of images based on their characteristics. In this work, we consider logos which come under image category for the purpose of classification. A logo is a symbolic representative of any industry or organization or institution, which symbolizes the functionality of their respective work.

---

\* Corresponding author. Tel.: +91-9632454934; fax: +0-000-000-0000 .
 *E-mail address:*vinaykumar.natraj@gmail.com

Once a logo is designed for any organization or a company, it needs to be tested for its originality and uniqueness. If not, many intruders can design logos which look very similar to the existing logos and may change the goodness of the company or organization. To avoid such trade infringement or duplication, a system which tests newly designed logos for its originality is required. To test for the originality, the system has to verify the newly designed logo by comparing with the existing logos. Since the number of logos available for comparison is huge, either a quick approach for comparison or any other alternative need to be investigated. One such alternative is to identify the class of logos to which the newly designed logo belongs and then verifying it by comparing with only the samples (logos) of its corresponding class. Thus, the process of classification followed by verification reduces the search space of a logo classification system to a greater extent. With this motivation, we have designed a logo classification system which classifies the logos based on their appearance.

*1.1. Related Work*

In literature, we can find many works on logo detection, logo extraction, logo retrieval, and logo classification.

Detection and Extraction of logos from documents is a major task involved in identifying the documents related to the particular company or organization[1]. In this direction, [2,3] presented an approach based on multi-scale boosting strategy for detection and extraction of logos from the documents. In [4], the authors have extracted the logos in two different stages. In the first stage, wavelet transform is used for logo extraction and in the second stage different thresholds are used for logo extraction. Later, KNN and MLP classifiers are used to classify logos. In [5], a system is proposed which handles document logo images and classifies the logos irrespective of geometric transformation. Even we can find a couple of works [6,7] related to logo classification in document images. Apart from logos detection, extraction and classification in document images, the detection of logos from the motor vehicles is also used for the purpose of motor product authentication. A rapid and robust method for vehicle logo detection is proposed in [8]. They applied directional filters and saliency map to highlight the vehicle logo region and locate the region. Vehicle logos can be detected rapidly without learning and also adaptable to different situations without adjusting the parameters.

Similarly, we can find many existing logo image retrieval systems [9,10,11,12], which can retrieve the similar logo images to a given query logo image. In [13], a logo retrieval system is proposed based on the contour information of logo images. They have used Fourier descriptors for representing the various shapes like rectangle, polygon, ellipse and B-spline. In [14], an efficient logo retrieval system is proposed using shape context descriptors. In [15], Zernike moments are used for describing the global and curvature features of a logo image. In [16], two different descriptors are combined for the effective retrieval of logo images. In [17], a retrieval system is proposed based on fuzzy color classification of logo images. In [18], a logo retrieval system based on wavelet co-occurrence histogram (WCH) is proposed.

Apart from the above works, there are couple of works available which work directly on the logo images. In [19], an attempt towards classifying the logos of the University of Maryland (UMD) logo database is made. Here, the logo images are classified as either degraded logo images or non-degraded logo images. In [12], a logo classification system is proposed for classifying the logo images captured through mobile phone cameras with a limited set of images. In [20], a comparative analysis of invariant schemes for logo classification is presented. From the literature, it can be observed that, in most of the works, the classification of logos has been done only on logos present in the document images. Also, there is no work available for classification of color logo images. Keeping this in mind, we thought of classifying the color logos based on their appearance [21].

In this paper, an approach based on various combinations of features for classification of color logos is proposed. Due to the lack of a huge color logo image dataset, a color logo image dataset consisting of 5044 logo images is created. Further, the logos are categorized as logos with fully text, logos with fully symbols and logos consisting of both text and symbol. Some of these color images are illustrated in figure 1. For the purpose of classification, the appearance based features such as color, texture and shape are extracted. For color feature extraction, the RGB image is divided into eight coarse partitions or blocks and finds the mean and percentage of individual block with respect to each color component (red / blue / green). For texture feature extraction, the image is processed using

Steerable Gaussian filter decomposition with four different orientations ($0^0$, $-45^0$, $+45^0$, $90^0$), then the mean and standard deviation at each decomposition is computed. For shape features, Zernike moments shape descriptor is used in two different orientations ($0^0$ and $90^0$) [22].

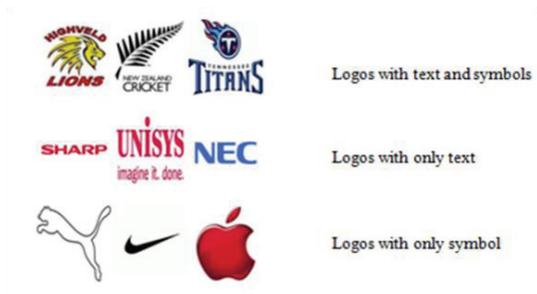

Fig. 1. Illustration of color logo images with logo images with both text and symbols, with only texts, and with only symbols.

The rest of the paper is organized as follows. In section 2, the details of the proposed logo classification system are explained. The experimental setup and detailed results are given in section 3. Further, we come up with the conclusion in section 4.

**2. Proposed Model**

Different steps involved in the proposed logo classification model are shown in figure 2. Our model is based on classifying a color logo as either a logo with only text or a logo with only symbols or a logo with both text and symbol. The different stages of the proposed model are explained in the following subsections.

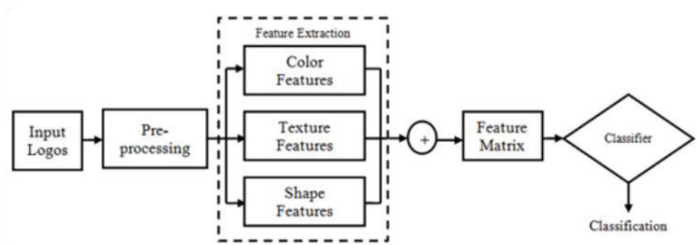

Fig. 2. Architecture of the proposed model.

*2.1. Pre-processing*

In this stage, we recommend two different pre-processing tasks, namely, image resizing and gray scale conversion. Initially, we resize all the logo images of dimension M x N into m x n to maintain the uniformity in the dimensions of the logo images. Then, we convert the RGB logo images into gray scale images. The conversion helps in extracting the texture and shape features from gray scale images [21].

*2.2. Feature Extraction*

In this work, we recommend three different appearance based (global) features namely, color, texture and shape features from an input color logo image. Color features are extracted by using eight partition method, texture features are extracted from steerable Gaussian filters and shape features are extracted from Zernike moments as follows.

These features are recommended for extraction from logo images, as the logos are very distinct in nature with respect to color, texture, and shape properties.

*2.2.1 Color Features*

In general, color is one of the most dominant and distinguishable global visual feature in describing an image. So, we have considered color as one of the three appearance based features for distinguishing different types of logos.

After resizing all logo images, we extract individual components of an RGB image separately as red component, green component and blue component images as shown in figure 3. Let $X^1$, $X^2$, and $X^3$ represent Red, Green, and Blue components respectively. Each component is then uniformly divided into 8 coarse partitions. The mean value of each partition is computed as its quantized color using equation 1. Similarly, the mean value of each partition for all three components is computed.

$$\overline{X}_i^k = \frac{\sum x_i^k}{No.\,of\ pixels\ in\ the\ respective\ partition} \qquad (1)$$

where $1 \leq i \leq 8, k = 1, 2, 3$

Now, we have 24 different scalar values representing mean values of each partition for all three components. These 24 values represent the quantized value of each partition of the respective R, G, and B components.

Further, we calculate the percentage of $i^{th}$ partition of the $k^{th}$ component using equations from (2) to (4). The calculation of percentage brings out the degree of each color component that is suitable in distinguishing among the images [21].

$$P_{1,i} = \left(\frac{\overline{X}_i^1}{TotalColor_i}\right) * 100 \qquad (2)$$

$$P_{2,i} = \left(\frac{\overline{X}_i^2}{TotalColor_i}\right) * 100 \qquad (3)$$

$$P_{3,i} = \left(\frac{\overline{X}_i^3}{TotalColor_i}\right) * 100 \qquad (4)$$

where, $TotalColor_i = \left(\overline{X}_i^1 + \overline{X}_i^2 + \overline{X}_i^3\right), 1 \leq i \leq 8$

In the above equations, $P_{1,i}, P_{2,i}, and\ P_{3,i}$ represent the percentages of each R, G, and B components respectively. So, we have obtained 24 different percentage values from eight different partitions of all three components. Finally, we combine the quantized values and percentage values to get a row vector consisting of 48 different scalar values for a logo image. Similarly, color features are extracted for all logo images.

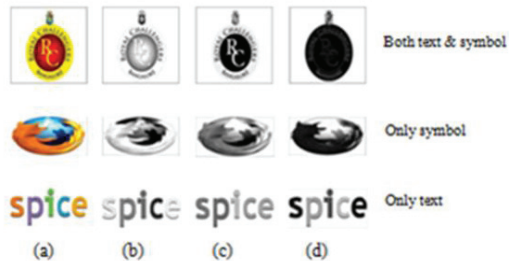

Fig. 3. Sample logo images of our logo dataset (a) Original logo image, (b) Red component of a logo, (c) Green Component of a logo, (c) Blue component of a logo.

*2.2.2   Texture Features*

Most of natural surfaces exhibit texture, which is an important low-level visual feature. Texture recognition will therefore be a natural part of many computer vision systems. In this work, we have used rotation-invariant and scale-invariant texture representatives based on steerable Gaussian filters [21].

For a given gray scale logo image, we have extracted eight different texture features as explained in [21]. These features are obtained from four different orientations viz., $0^0$, $90^0$, $45^0$, and $-45^0$ which preserve the textural appearance of a given logo image as shown in figure 4.

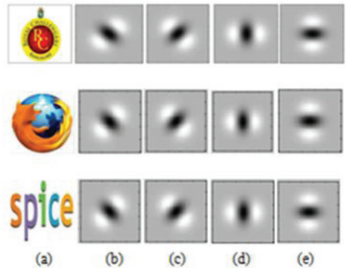

Fig. 4. Texture features at different orientations, (a) Original logo image, (b) The texture image at $-45^0$, (c) The texture image at $45^0$, (d) The texture image at $90^0$, (e) The texture image at $0^0$.

*2.2.3   Shape Features*

A shape plays a major role in recognition and perception of objects. Objects shape features provide a powerful a clue to object identity [9].

Here, we have used pseudo-Zernike moments as shape descriptor, which is invariant to all geometrical transformations and more robust to noise [22, 23, 24]. It computes mainly two parameters like Amplitude and phase angle both in horizontal and vertical directions of a logo image. More details on the theory of Zernike moment feature extraction process can be found in [25].

Here, we get totally four different shape features viz., amplitude and phase angle with respect to vertical direction and the same with respect to horizontal direction. These features describe a logo image. These four shape features are further combined with texture and color features for effective classification of logo images.

*2.3. Feature Level Fusion*

In this work, we have considered fusion at feature level. It helps us in classifying the objects by considering a set of fused features. With the use of fusion of features, the classification should be done more accurately than the object classified with a single feature [26, 27]. In this regard, we have considered feature level fusion for classifying the logos.

If we consider various combinations of features for fusion then there exists *n* features and t combinations (where t<=n). The various feature combinations obtained are $C_1^n, C_1^n, ..., C_t^n$. These combinations of features are used for classification of color logo images. In our work totally there are three different features color, texture, and shape features. These three features are considered for computing the various combinations of features for classification of color logo images.

The three features which are extracted for classification can be seen at different perception levels. So, these features are normally not normalized. To normalize these features, a normalization technique is used and is given in equation 5. After normalization, the value of each feature ranges in between 0 and 1.

$$F_j = \frac{x_{ij}}{\max(x_{ij})} \tag{5}$$

Where, i=1 to number of samples; and j=1 to number of features

*2.4. Classification*

In this work, we adopted the K-Nearest Neighbour classifier for the purpose of classification.

In, KNN classifier, the features of a test logo image is compared against the features of all referenced logo images. The Euclidean distance is used for the computation of distances between the features of a test logo image and all reference logo images. The class label of a reference logo image is assigned to a test logo image, if the distance between a test logo image and a reference logo image is minimum among all the distances. Thus a test logo image is classified as a member of any one of the three classes. The same procedure is followed in classifying all the remaining logo images.

## 3. Experimentation

*3.1. Dataset*

For experimentation, we have created our own dataset named "UoMLogo database" consisting of 5044 color logo images. The "UoMLogo Database" mainly consists of color logo images of different universities, brands, sports, banks, insurance, cars, and industries etc. which are collected from the internet. This dataset mainly categorized into three classes, BOTH logo image (a combination of TEXT and SYMBOL), TEXT logo image, SYMBOL image. Within class, there exist ten different subclasses. Figure 5 shows the sample images of the UoMLogo Dataset. The complete details of the created database are given table 1.

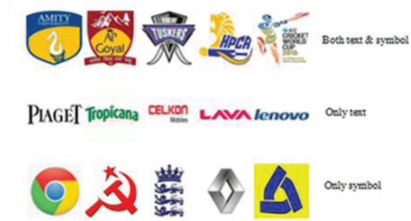

Fig. 5. Samples logo images of UoMLogo Dataset.

Table 1. Details UoM logo dataset.

| Class | University | Sports | Banks | Insurance | Cars | Brands | Govt. & Political Party | U N O | Media | Industry | Total |
|---|---|---|---|---|---|---|---|---|---|---|---|
| BOTH | 929 | 1462 | 21 | 128 | 33 | 155 | 42 | 171 | - | 230 | 3171 |
| TEXT | - | 28 | - | 35 | 08 | 538 | - | - | 358 | 279 | 1246 |
| SYMBOL | - | 139 | 209 | 25 | 70 | 25 | 108 | 28 | - | 23 | 627 |
| TOTAL | 929 | 1629 | 230 | 188 | 111 | 718 | 150 | 199 | 358 | 532 | 5044 |

*3.2. Experimental setup*

In preprocessing step, for the sake of simplicity and uniformity in extracting the features from a color logo image, we have resized every image into 200 x 200 dimensions. To extract texture and shape features, the color logo images are converted into grayscale images, as these two features are independent of color features. In feature extraction stage, three different features mainly: color, texture and shape features are extracted. These features are extracted from logo images as discussed in the section 2.2. Further, these features are normalized and used for classification purpose. The classification is done based on the various combinations obtained from these three features. Therefore, the various combinations obtained for classification are $C_1^3, C_2^3, and\ C_3^3$. Here, $C_1^3$ produces with three individual features (color(c), texture(t), and shape(s)) separately for classification. Next, $C_2^3$ produces with three pair of features (c+t, c+s, and t+s) for classification. At last, $C_3^3$ produces with a single combination (c+t+s), consists of all three features for classification. Totally, these seven combinations are used for classification of color logo images.

KNN classifier is used for classification purpose and the value of K is set to 1 in all the experimentation conducted.

In our proposed classification system, the dataset is divided into training and testing. Seven sets of experiments have been conducted under varying number of training set images as 20%, 30%, 40%, 50%, 60%, 70% and 80%. While testing stage, the system uses remaining 80%, 70%, 60%, 50%, 40%, 30%, and 20% of logo images respectively for classifying them as any one of the three classes. At each testing stage, the classification results are presented by the confusion matrix. The performance of the classification system is evaluated using classification accuracy, precision, recall, and F-Measure computed from the confusion matrix [28].

### 3.3. Experimental results

The performance of the proposed classification system is evaluated using classification accuracy, precision, recall and F-Measure computed from the confusion matrix.

Let us consider a confusion matrix $CM_{ij}$, generated during classification of color logo images at some testing stage. From this confusion matrix, the accuracy, the precision, the recall, and the F-Measure are all computed to measure the efficacy of the proposed logo image classification system. The overall accuracy of a system is given by:

$$Accuracy = \frac{No.\,of\,correctly\,classified\,samples}{Total\,number\,of\,samples} * 100 \quad (6)$$

The precision and recall can be computed in two ways. Initially, they are computed with respect to each class and later with respect to overall classification system. The class wise precision and class wise recall is computed from the confusion matrix are given in equations (7) and (8) respectively.

$$P_i = \frac{No.\,of\,correctly\,classified\,samples}{No.\,of\,samples\,classified\,as\,a\,member\,of\,a\,class} * 100 \quad (7)$$

$$R_i = \frac{No.\,of\,correctly\,classified\,samples}{Expected\,number\,of\,samples\,to\,be\,classified\,as\,a\,member\,of\,a\,class} * 100 \quad (8)$$

Where, i=1,2,…,n; n=No. of Classes

The system precision and system recall computed from the class wise precision and class wise recall is given by:

$$Precision = \frac{\sum_{i=1}^{n} P_i}{n} \quad (9)$$

$$Recall = \frac{\sum_{i=1}^{n} R_i}{n} \quad (10)$$

The F-measure computed from the precision and recall is given by:

$$F-Measure = \frac{2*Precision*Recall}{Precision+Recall} \quad (11)$$

The classification results obtained for different training and testing percentage of samples are tabulated. Tables from 2 to 8 show the overall accuracy, precision, recall, and F-measure obtained from the classification system by taking into account of various combinations of features (varied from 1 to 3). Here, precision and recall are computed from the results obtained from the class wise precision and class wise recall respectively.

Table 2. Classification Accuracy, Precision, Recall and F-Measure obtained for varying training and testing percentage using K-NN classifier (Color Features).

| Train-Test % | Accuracy | Precision | Recall | F-Measure |
|---|---|---|---|---|
| 20-80 | 55.29 | 41.86 | 42.88 | 42.36 |
| 30-40 | 54.52 | 41.21 | 42.16 | 41.68 |
| 50-50 | 52.30 | 39.07 | 40.28 | 39.67 |
| 60-40 | 51.29 | 39.18 | 40.29 | 39.73 |
| 70-30 | 48.16 | 37.70 | 38.67 | 38.18 |
| 80-20 | 42.39 | 33.77 | 33.95 | 33.86 |

Table 3. Classification Accuracy, Precision, Recall and F-Measure obtained for varying training and testing percentage using K-NN classifier (Texture Features).

| Train-Test % | Accuracy | Precision | Recall | F-Measure |
|---|---|---|---|---|
| 20-80 | 53.19 | 39.46 | 39.52 | 39.49 |
| 30-40 | 53.16 | 41.15 | 41.49 | 41.32 |
| 50-50 | 53.26 | 40.90 | 41.18 | 41.04 |
| 60-40 | 52.32 | 40.92 | 41.37 | 41.15 |
| 70-30 | 51.39 | 41.80 | 42.47 | 42.13 |
| 80-20 | 47.75 | 40.34 | 40.55 | 40.45 |

Table 4. Classification Accuracy, Precision, Recall and F-Measure obtained for varying training and testing percentage using K-NN classifier (Shape Feature).

| Train-Test % | Accuracy | Precision | Recall | F-Measure |
|---|---|---|---|---|
| 20-80 | 54.48 | 33.58 | 33.15 | 33.37 |
| 30-40 | 55.28 | 33.49 | 33.18 | 33.34 |
| 50-50 | 56.07 | 33.15 | 33.04 | 33.09 |
| 60-40 | 56.96 | 34.82 | 33.92 | 34.36 |
| 70-30 | 58.09 | 36.14 | 34.58 | 35.34 |
| 80-20 | 57.41 | 34.36 | 33.93 | 34.14 |

Table 5. Classification Accuracy, Precision, Recall and F-Measure obtained for varying training and testing percentage using K-NN classifier (Color+Texture Features).

| Train-Test % | Accuracy | Precision | Recall | F-Measure |
|---|---|---|---|---|
| 20-80 | 56.81 | 41.87 | 43.10 | 42.48 |
| 30-40 | 55.74 | 42.28 | 43.14 | 42.71 |
| 50-50 | 55.54 | 43.17 | 43.77 | 43.47 |
| 60-40 | 52.40 | 40.80 | 41.66 | 41.23 |
| 70-30 | 48.81 | 38.81 | 39.54 | 39.17 |
| 80-20 | 42.06 | 34.05 | 34.16 | 34.11 |

Table 6. Classification Accuracy, Precision, Recall and F-Measure obtained for varying training and testing percentage using K-NN classifier (Color+Shape Features).

| Train-Test % | Accuracy | Precision | Recall | F-Measure |
|---|---|---|---|---|
| 20-80 | 56.06 | 42.03 | 43.22 | 42.62 |
| 30-40 | 54.92 | 40.81 | 41.82 | 41.31 |
| 50-50 | 53.12 | 39.94 | 40.93 | 40.43 |
| 60-40 | 52.04 | 40.00 | 40.89 | 40.44 |
| 70-30 | 48.86 | 39.00 | 39.71 | 39.35 |
| 80-20 | 43.58 | 35.69 | 35.75 | 35.72 |

Table 7. Classification Accuracy, Precision, Recall and F-Measure obtained for varying training and testing percentage using K-NN classifier (Shape+Texture Features).

| Train-Test % | Accuracy | Precision | Recall | F-Measure |
|---|---|---|---|---|
| 20-80 | 54.00 | 40.53 | 40.47 | 40.50 |
| 30-40 | 54.49 | 42.12 | 42.54 | 42.33 |
| 50-50 | 54.61 | 42.54 | 42.94 | 42.74 |
| 60-40 | 53.43 | 42.78 | 43.53 | 43.15 |
| 70-30 | 51.64 | 42.94 | 44.07 | 43.50 |
| 80-20 | 47.88 | 41.13 | 42.19 | 41.65 |

Table 8. Classification Accuracy, Precision, Recall and F-Measure obtained for varying training and testing percentage using K-NN classifier (Color+Texture+Shape Features).

| Train-Test % | Accuracy | Precision | Recall | F-Measure |
|---|---|---|---|---|
| 20-80 | 63.01 | 44.99 | 43.62 | 44.29 |
| 30-40 | 62.85 | 47.39 | 44.82 | 46.07 |
| 50-50 | 62.52 | 51.12 | 44.79 | 47.75 |
| 60-40 | 59.69 | 45.62 | 43.48 | 44.52 |
| 70-30 | 56.02 | 46.96 | 42.51 | 44.62 |
| 80-20 | 48.98 | 40.47 | 36.92 | 38.62 |

From tables 2 to 8, it is observed that the classification of logo images yields good results for lesser percentage of training samples. The results gradually decrease as the number of training samples increase. This is due to the large variation of size with respect to increase in the size of the training samples present in the respective three classes (see Table 1). Also, from the above tables, it is clearly observed that, the combination of features results with maximum accuracy, maximum precision, maximum recall and maximum F-Measure. These best results are all belongs to the feature combinations considered with more than one feature at a time. So, the best classification results in terms of accuracy, precision, recall and F-measure obtained for respective combination of features are tabulated in table 9.

Table 9. Best Accuracy, Precision, Recall and F-Measure obtained for respective combination of features under varied train-test percentage.

| Train-Test % | Accuracy | Precision | Recall | F-Measure |
|---|---|---|---|---|
| 20-80 | 63.01 (c+t+s) | 44.99 (c+t+s) | 43.62 (c+t+s) | 44.29 (c+t+s) |
| 30-40 | 62.85 (c+t+s) | 47.39 (c+t+s) | 44.82 (c+t+s) | 46.07 (c+t+s) |
| 50-50 | 62.52 (c+t+s) | 51.12 (c+t+s) | 44.79 (c+t+s) | 47.75 (c+t+s) |
| 60-40 | 59.69 (c+t+s) | 45.62 (c+t+s) | 43.48 (c+t+s) | 44.52 (c+t+s) |
| 70-30 | 56.02 (c+t+s) | 46.96 (c+t+s) | 42.51 (c+t+s) | 44.62 (c+t+s) |
| 80-20 | 47.88 (t+s) | 41.13 (t+s) | 42.19 (t+s) | 41.65 (t+s) |

From the above discussion, it is very very clear that the color logo image classification system yields promising results with consideration of more than one feature at a time.

## 4. Conclusion

In this paper an approach in classifying the color logo images into the pre-defined three classes is proposed. In classifying a logo image, the global characteristics of logo images are extracted. The various combinations of these characteristics are explored for a color logo image classification. Further, K-NN classifier is used for classification. The effectiveness of the proposed classification system is validated through well known measures like accuracy, precision, recall and F-Measure. Finally, the paper concludes with an understanding that the promising classification results are obtained only for the feature combination considered more than one at a time.

## Acknowledgements

The work by N. Vinay Kumar was financially supported by DST-INSPIRE Fellowship under Department of Science and Technology, Government of India, India.